\documentclass[fleqn,10pt]{wlscirep}
\usepackage[utf8]{inputenc}
\usepackage[T1]{fontenc}
\usepackage{lineno}
\usepackage{float}
\usepackage{hyperref}
\restylefloat{table}

\title{The "LLM World of Words" English free association norms generated by large language models}

\author[1,*]{Katherine Abramski}
\author[2]{Riccardo Improta}
\author[3,+]{Giulio Rossetti}
\author[2,+]{Massimo Stella}
\affil[1]{University of Pisa, Department of Computer Science, Pisa, Italy}
\affil[2]{University of Trento, Department of Psychology and Cognitive Science, Trento, Italy}
\affil[3]{National Research Council of Italy, Institute of Information Science and Technologies, Pisa, Italy}
\affil[*]{Corresponding author: Katherine Abramski (katherine.abramski@phd.unipi.it).}
\affil[+]{These authors contributed equally}

\begin{abstract}
Free associations have been extensively used in cognitive psychology and linguistics for studying how conceptual knowledge is organized. Recently, the potential of applying a similar approach for investigating the knowledge encoded in LLMs has emerged, specifically as a method for investigating LLM biases. However, the absence of large-scale LLM-generated free association norms that are comparable with human-generated norms is an obstacle to this new research direction. To address this limitation, we create a new dataset of LLM-generated free association norms modeled after the "Small World of Words" (SWOW) human-generated norms consisting of approximately 12,000 cue words. We prompt three LLMs, namely Mistral, Llama3, and Haiku, with the same cues as those in the SWOW norms to generate three novel comparable datasets, the "LLM World of Words" (LWOW). Using both SWOW and LWOW norms, we construct cognitive network models of semantic memory that represent the conceptual knowledge possessed by humans and LLMs. We demonstrate how these datasets can be used for investigating implicit biases in humans and LLMs, such as the harmful gender stereotypes that are prevalent both in society and LLM outputs.

\end{abstract}
\begin{document}

\flushbottom
\maketitle

\thispagestyle{empty}

\section*{Background \& Summary}

How is conceptual knowledge organized in the mind? Such a question has long been the focus of linguists and cognitive psychologists who aim to better understand the human language capacity \cite{de2013better,kenett2017semantic}. Recently, this question has become increasingly relevant in the field of artificial intelligence, particularly regarding large language models (LLMs). Human semantic memory -- the repository of conceptual knowledge that encompasses how words get their meaning \cite{davisaitchison} -- forms the foundation of human language and thought, and thus, its structure and properties influence how we reason, form beliefs and make decisions \cite{vankrunkelsven2018predicting,samuel2023predicting}, ultimately shaping our social and political systems. Similarly, the semantic representations that comprise the knowledge encoded in LLMs are the underlying source behind the outputs they produce, and as LLMs become more integrated into our everyday lives, these outputs have an increasing impact on society \cite{shiffrin2023probing,binz2023using}. Thus, the study of the structure and properties of semantic memory is central to understanding not only our own thinking and reasoning, but also the "thinking" and "reasoning" of LLMs, which carries important societal implications.

Studying semantic memory involves creating representations of word meanings (semantic representations), often in terms of how words relate to other words. In humans, one common way to do this is using free associations \cite{steyvers2005large,de2013better,stella2019forma}, which are usually accessed by prompting participants with a cue word and asking them to come up with (typically three) associated responses. Since the task is context neutral, responses represent the associative knowledge of words that we possess at an implicit level. Free associations have been extensively used in cognitive psychology and linguistics for studying lexical retrieval \cite{de2013better, vankrunkelsven2018predicting}, semantic organization \cite{steyvers2005large}, and similarity judgments \cite{de2008word, kenett2017semantic, de2019small}. They have also been used for studying differences in cognitive processing between concrete and abstract words, i.e. concreteness effects \cite{hill2014quantitative
}. Given that free associations have been shown to correlate with stable implicit attitudes \cite{schnabel2013free}, they have also been used for studying affective biases \cite{stella2019forma}. Investigations of conceptual knowledge using free associations are often conducted within network models of semantic memory built from free associations by connecting cue words to their responses. This results in a complex network structure of human conceptual knowledge in which words get their meanings through relationships to other concepts. Such models enable the investigation of complex cognitive processes that take place within semantic memory. In fact, cognitive network models of semantic memory have been used to gain powerful insights about a variety of human cognitive phenomena such as language learning, \cite{citraro2023feature, steyvers2005large}, creativity \cite{kenett2016examining, beaty2023associative, benedek2017semantic}, personality traits like openness to experience \cite{samuel2023predicting}, and autism spectrum disorder \cite{kenett2016hyper}.

While free associations have been widely used for studying semantic memory in humans, very different approaches have been applied for investigating conceptual knowledge in language models. Typically, semantic representations are directly accessed from the model's embedding space in the form of word embeddings \cite{apidianaki2022word}, i.e. vector representations of words whose meanings are derived from statistical relationships in the training data. Word embeddings provide an advantageous way for investigating certain aspects of semantic memory because of the types of mathematical operations that can be applied within the embedding space. Specifically, the semantic similarity between two words can easily be calculated by computing the cosine similarity between their word vectors. Thus, this approach can be used for extracting word associations directly from the model's architecture, and these associations can be used to study several aspects related to the model's conceptual knowledge \cite{rodriguez2020word, yao2022wordties}. In recent years, there has been a great interest in investigating the biases encoded in language models, and measuring the strength of associations between words within the model's embedding space has been extensively applied for accomplishing this goal \cite{bolukbasi2016man, kurita2019measuring, manzini2019black, caliskan2017semantics, bommasani2020interpreting}. Such an approach follows the same idea that the Implicit Association Test (IAT) \cite{greenwald1998measuring} uses to investigate implicit attitudes in humans. Essentially, it involves using the cosine similarity to measure the strength of the association between pairs of words for assessing biases, for example, \textit{man -- doctor} and \textit{woman -- nurse}. Depending on the strength of these respective associations, this method can reveal certain biases encoded in the model's embedding space, such as gender biases that reflect the implicit analogy, \textit{man is to doctor as woman is to nurse} \cite{bolukbasi2016man}. 

Extracting associations from the embedding space for investigating the conceptual knowledge encoded in language models has many advantages, but it also has some important limitations. Perhaps the most significant limitation is that this approach works well for older language models that use static word embeddings -- which represent word meanings at the type level -- but not so well for newer models, i.e. LLMs, that use contextual word embeddings -- which represent word meanings at the token level. Operations on contextual embeddings require that the embeddings are first transformed into static embeddings \cite{bommasani2020interpreting}, but this can introduce bias and distort similarity estimates \cite{apidianaki2022word}. Another downside to the approach of accessing the embedding space is that it limits the possibility to make comparisons across models and with humans, since the cognitive architecture of each model is vastly different. These limitations have led to a recent shift from the bottom-up approach of accessing the embedding space for investigating the knowledge encoded in language models, towards a top-down approach that involves prompting models with tasks and using their output to make inferences about the knowledge encoded in their embedding space \cite{binz2023using, srivastava2022beyond, shiffrin2023probing}. This approach mirrors methods from cognitive psychology that use behavioral experiments to make inferences about the workings of the human mind, and thus it has been aptly named "machine psychology" \cite{hagendorff2023machine}. One of the main advantages of machine psychology is that it allows for the use of preexisting and well-studied tools, measures, and methodologies that have been applied to humans for years \cite{binz2023using}. Thus, rather than developing completely new methodologies, the challenge of machine psychology lies in adapting the existing methodologies from cognitive science and psycholinguistics so that they can be applied to LLMs to gain insights about specific cognitive phenomena. Recent cutting edge studies have applied the machine psychology approach to investigate the semantic capabilities of LLMs compared to humans \cite{wang2024fluency, suresh2023conceptual, digutsch2023overlap}, demonstrating that this is a promising direction of research.

Up until this point, we have touched upon three main ideas:
\begin{enumerate}
    \item The use of free associations for investigating psychological phenomena in humans has long-standing relevance \cite{steyvers2005large};
    \item There has been extensive work \cite{bolukbasi2016man}, as well as a great interest, in using word associations extracted from the embedding spaces of language models for investigating biases encoded in their architecture;
    \item There has been a recent shift from the bottom-up approach to a top-down machine psychology approach for investigating the knowledge encoded in LLMs \cite{binz2023using,abramski2023cognitive}.
\end{enumerate}

\noindent Together, these ideas point to the need for a new direction of research that uses LLM-generated free associations in order to investigate the structure and properties of conceptual knowledge in LLMs. While there exist several datasets of free association norms generated by humans \cite{de2019small,nelson2004university,wilson1988eat}, to the best of our knowledge, there are no openly available datasets of LLM-generated free association norms that are comparable in scale and breadth to existing human-generated norms. 

To fill this research gap, we present the LLM World of Words (LWOW), a dataset of English free association norms including millions of responses generated by three different LLMs: Mistral, Llama3, and Haiku. LWOW is modeled after the largest dataset of human-generated English free association norms, called the Small World of Words (SWOW) \cite{de2019small}, which has been used extensively in many psychological and linguistic studies \cite{samuel2023predicting}. LWOW contains over 12,000 cue words, each with 3 responses, repeated 100 times, for a total of over 3 million responses. The LWOW dataset is generated using the same cue words used in SWOW, following the same methodology, though instead of asking humans to produce responses to the cues, we asked the three LLMs to produce responses to the cues. The result is three sets of LLM-generated free association norms that are directly comparable to the SWOW dataset. The LWOW dataset, combined with the original SWOW dataset, will enable investigations of the structure and properties of conceptual knowledge in both humans and LLMs, allowing for unprecedented comparisons between the two. 

We anticipate that LWOW will be particularly useful for investigating the nature of implicit biases in LLMs as they relate to human biases, such as the gender and racial stereotypes that are prevalent both in society \cite{manzini2019black} and in LLM outputs \cite{hagendorff2023machine}. For this reason, as part of the validation and usage notes of this dataset, we construct cognitive network models of semantic memory \cite{stella2019forma} from both the SWOW and LWOW datasets (humans and LLMs) and we use them to investigate the presence of gender stereotypes within their associative knowledge structures.

The remainder of this paper is structured as follows. In \textit{Methods}, we describe the methodology used to (1) generate the datasets, (2) preprocess the data, and (3) build network models of semantic memory from the data. In this section, we also provide summary statistics of the datasets, the network models, and comparisons between the networks. In \textit{Data Records}, we provide a detailed description of the repository containing the code and original datasets used to generate the data. In \textit{Data Validation}, we demonstrate the validity of the data by simulating the cognitive mechanisms that underlie
semantic priming within the network models, showing that activation patterns within the networks correlate with behavioral data from a well-known psycholinguistic experiment, i.e. the lexical decision task (LDT). In \textit{Usage Notes}, we show how the LWOW datasets can be used for investigating biases in humans and LLMs. Specifically, we adapt the methodology used in \textit{Data Validation} to measure the strength of association between pairs of words in order to quantify gender biases within the network models. Finally, in \textit{Code Availability} we provide details on how to access the code in order to reproduce the analyses.


\section*{Methods}


\subsection*{Data generation}
Since the LWOW datasets are based off the Small World of Words (SWOW) human-generated norms, we first gathered the cue words from the original SWOW dataset. The SWOW dataset was downloaded from \textit{https://smallworldofwords.org/en/project/}. We used the preprocessed data (SWOW-EN.R100.csv) with 12,282 cue words and 100 sets of responses per cue. We used a list of these cues as input to the LLMs along with a prompt that aimed to mimic the instructions provided to humans in the original SWOW free association task. The following prompt was given to the LLMs: 

\begin{figure}[h!]
\centering
\includegraphics[width=.9\linewidth]{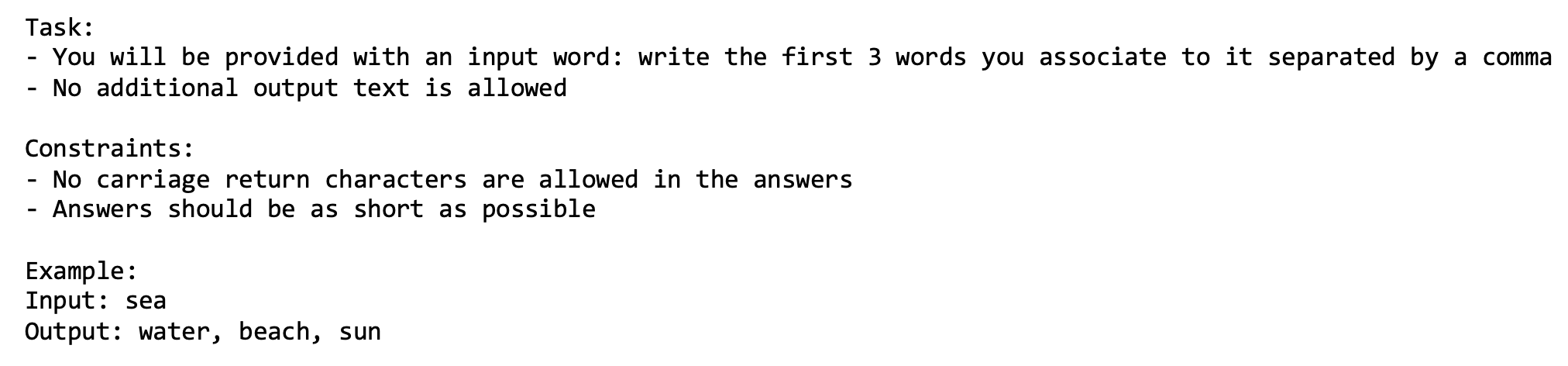}
\end{figure}

\noindent This prompt was repeated 100 times for each cue word in order to generate a dataset with the same number of responses as the original (preprocessed) SWOW dataset. In what follows, we describe how the LLM-generated output as well as the original SWOW output were further processed.

\subsection*{Data processing}
The original SWOW data \cite{de2019small} were already preprocessed, but in order to facilitate analyses and data alignment, we applied additional data preprocessing to both the original SWOW data and the output produced by all three language models. From this point on, we refer to the SWOW dataset, including all subsequent modifications, as the Human dataset. The preprocessing steps applied to all four datasets, which were done in python, are as follows. First, all cues and responses were made lowercase. Then, the articles \textit{a, an, the}, and the preposition \textit{to} were removed from the beginning of responses unless they were among the original cues (e.g. \textit{a lot}). Some responses included underscores, and these were replaced with spaces. Also, some responses incorrectly lacked spaces or hyphens (e.g. \textit{throwout, checkin}). In order to ensure that these responses were not excluded in later analyses, we created a mapping dictionary using WordNet \cite{miller1995wordnet} (implemented in the python library \textit{nltk}) to resolve this issue. Specifically, we took all the words in WordNet \cite{miller1995wordnet} that have either spaces or hyphens (e.g. \textit{throw out, check-in}) and we removed them to create a one-to-one mapping to correct these errors in the responses. Spelling corrections were also applied to both cues and responses according to a dictionary that was used to process the original SWOW data. The spelling dictionary included the correction of commonly misspelled words (e.g. \textit{recieve} to \textit{receive}) but it also mapped British spelling to American spelling (e.g. \textit{colour} to \textit{color}). Next, cues and responses were lemmatized using WordNet's lemmatizer, changing plural nouns to singular nouns (e.g. \textit{men} to \textit{man}) but leaving tensed verbs unchanged (e.g. \textit{cooking}, \textit{determined}). Next, we added or removed data to ensure exactly 100 repetitions per cue. This step was needed for the Human data because the spelling corrections and the lemmatization of cues resulted in more than 100 repetitions for some cues, while some LLMs, namely Llama3, failed to generate 100 repetitions per cue. Thus, we ensured 100 repetitions per cue by adding blank responses when there were less than 100 repetitions for a cue, while sampling randomly when there were more than 100 repetitions per cue. Finally, responses that were identical to their corresponding cues were removed, and duplicate responses within the same set of three responses were removed. After this preprocessing procedure, each dataset resulted in a total of 11,545 cues, compared to the 12,282 cues in the original SWOW dataset. Table 
\ref{tab:data_stats} shows the statistics of each dataset, including the number of cues, number of total responses, number of unique responses, and percentage of missing responses. The Human dataset has the most unique responses, closely followed by Llama3. Mistral and Haiku have far fewer unique responses.

\begin{table}[H]
\centering
\begin{tabular}{ccccc}
\hline
\textbf{Network} & \textbf{Unique cues} & \textbf{Total responses} & \textbf{Unique responses} & \textbf{Missing responses} \\ \hline
Humans           & 11,545               & 3,148,578                & 116,640                   & 9.1\%                      \\
Mistral          & 11,545               & 3,268,206                & 41,369                    & 5.6\%                      \\
Llama3           & 11,545               & 3,348,049                & 105,367                   & 3.3\%                      \\
Haiku            & 11,545               & 3,403,644                & 15,275                    & 1.7\%                      \\ \hline
\end{tabular}

\caption{\textbf{Dataset statistics.} Cue and response statistics for all datasets after preprocessing. All networks have the same unique cues, but different numbers of total responses and unique responses. The Human network has the largest percentage of missing responses, but also the largest number of unique responses compared to all LLMs.}
\label{tab:data_stats}

\end{table}

While these data have undergone important preprocessing, this procedure did nothing to remove nonsensical responses, from both the Human and LLM datasets. Such responses are unsuitable for certain analyses, and we suggest additional cleaning procedures be applied to remove nonsensical responses, however, we include all these responses in the preprocessed data for several reasons. First, what can be considered a valid response is rather subjective. For example, while the human-generated response \textit{accidentally pressed enter instead of no more associations} is clearly not a response to the cue, some other responses such as \textit{violence against women} or \textit{easy to get along with} may be considered valid, despite not being words per se. Secondly, the types of invalid or nonsensical responses produced by humans and LLMs alike may be of great interest to some researchers. Thirdly, there is no systematic way to remove all nonsensical responses. Instead, identifying responses that are clearly nonsensical, such as \textit{printassociatedwordsinputword} requires applying a series of ad-hoc filters. We experimented with applying such filters, and we found that while some nonsensical responses are easily removed by searching for certain sets of strings, other invalid responses are harder to classify. For example, in the data generated by Mistral, the responses \textit{output}, \textit{input}, and \textit{association} appeared much more frequently than in the human data, and most are clearly invalid responses. However, such responses are difficult to identify because \textit{input} could be a perfectly valid response to cues such as \textit{output}, \textit{feedback}, or \textit{function}, but it is most likely an invalid response to cues like \textit{flower} or \textit{monkey}. For these reasons, we have not applied any filters to remove or correct these responses in the preprocessed data. Instead, we applied filters in the network building process to remove responses that, for the purpose of our analyses, we considered invalid.

\subsection*{Network construction}

As discussed earlier, free associations are often used to build network models of semantic memory. In this way, semantic memory is considered a complex system \cite{kenett2017semantic,kenett2016examining,stella2019forma}, and it can be studied as such even in case of systems without an explicit semantic memory, like LLMs, that are nonetheless capable of processing language \cite{abramski2023cognitive}. The network structure provides a quantitative framework within which certain cognitive phenomena can be investigated using the tools of network science. From the preprocessed data, we built network models of semantic memory for humans and all three LLMs by connecting cue words to their responses. The weight of the edge reflects the frequency of the response, so if \textit{cat} appears 20 times as the response to the cue \textit{dog}, then a directed edge of weight 20 is created from \textit{dog} to \textit{cat}. The networks are naturally directed, but they are transformed into undirected networks to facilitate the analyses that we will discuss in the next section. In cases in which there is a bidirectional edge, the largest of the two edge weights is maintained. So if the edge from \textit{dog} to \textit{cat} has a weight of 20 and the edge from \textit{cat} to \textit{dog} has a weight of 25, the undirected edge between \textit{cat} and \textit{dog} has a weight of 25. The full undirected networks are then filtered to remove unwanted nodes and edges. This is done first by removing nodes that are not in WordNet, and then by removing idiosyncratic edges, i.e. edges with a weight = 1, and finally, taking the largest connected component. This network filtering has a few advantages. First, the WordNet filter provides a standardized way to eliminate nonsensical and uncommon responses, since words in WordNet are at least English words, and the removal of idiosyncratic edges ensures that the association is something shared among two or more people (iterations in the case of the LLMs) and not just a fluke. A final advantage is that this filtering reduces the number of nodes and edges, making the networks more computationally manageable.

Table \ref{tab:net_stats} shows the network statistics for both the full networks (before filtering) and the reduced networks (after filtering) for each of the four datasets (Humans, Mistral, Llama3, and Haiku). Statistics include the number of nodes, number of edges, the network density, and the average degree. The reduced networks are the final versions of the networks, and from this point on, when we discuss the networks, we refer to the reduced networks. Among the reduced  networks, the largest network is Llama3 followed by Humans, Mistral, and finally Haiku. To assess the extent to which each LLM network is similar/different to the Human network, we made pairwise comparisons of nodes and edges between each LLM network and the Human network. The pairwise comparison statistics are shown in Table \ref{tab:net_comp}. For the node comparisons, we calculated the percentage of Human nodes not in the LLM network, the percentage of all nodes common to both networks, and the percentage of LLM nodes not in the Human network. For the edge comparisons, the same statistics were calculated, however they were calculated on the subgraphs of the node intersections of each pair of graphs.

\begin{table}[H]
\centering
\begin{tabular}{lccccc}
\hline
\textbf{Full Networks}    & \textbf{Network} & \textbf{Nodes} & \textbf{Edges} & \textbf{Density} & \textbf{Average degree} \\ \hline
                          & Humans           & 116,640        & 1,164,026      & 0.0002           & 20.0                    \\
                          & Mistral          & 42,073         & 417,697        & 0.0005           & 19.9                    \\
                          & Llama3           & 105,777        & 770,458        & 0.0001           & 14.6                    \\
                          & Haiku            & 17,679         & 77,698         & 0.0005           & 8.8                     \\
                          &                  &                &                &                  &                         \\ \hline
\textbf{Reduced Networks} & \textbf{Network} & \textbf{Nodes} & \textbf{Edges} & \textbf{Density} & \textbf{Average degree} \\ \hline
                          & Humans           & 24,308         & 317,344        & 0.0011           & 26.1                    \\
                          & Mistral          & 20,339         & 199,103        & 0.0010           & 19.6                    \\
                          & Llama3           & 38,987         & 546,866        & 0.0007           & 28.1                    \\
                          & Haiku            & 15,596         & 64,599         & 0.0005           & 8.3                     \\ \hline
\end{tabular}
\caption{\textbf{Network statistics.} The full networks are built using all cues and responses from the cleaned datasets, while the reduced networks are filtered by removing nodes not in WordNet, removing idiosyncratic edges, and then taking the largest connected component. The purpose of this network filtering is to remove senseless responses and to make the networks more computationally manageable.}
\label{tab:net_stats}
\end{table}


\begin{table}[H]
\centering
\begin{tabular}{ccccc}
\hline
\textbf{Nodes}       & \textbf{Comparison with Humans}                 & \textbf{Mistral}     & \textbf{Llama3}      & \textbf{Haiku}       \\ \hline
                     & Percentage of Human nodes not in LLM network    & 32\%                 & 16\%                 & 41\%                 \\
                     & Percentage of all nodes common to both networks & 59\%                 & 48\%                 & 57\%                 \\
                     & Percentage of LLM nodes not in Human network    & 19\%                 & 47\%                 & 8\%                  \\
\multicolumn{1}{l}{} & \multicolumn{1}{l}{}                            & \multicolumn{1}{l}{} & \multicolumn{1}{l}{} & \multicolumn{1}{l}{} \\ \hline
\textbf{Edges}       & \textbf{Comparison with Humans}                 & \textbf{Mistral}     & \textbf{Llama3}      & \textbf{Haiku}       \\ \hline
                     & Percentage of Human edges not in LLM network    & 69\%                 & 70\%                 & 84\%                 \\
                     & Percentage of all edges common to both networks & 23\%                 & 13\%                 & 15\%                 \\
                     & Percentage of LLM edges not in Human network    & 51\%                 & 81\%                 & 24\%                 \\ \hline
\end{tabular}

\caption{\textbf{Network comparisons.} The table shows pairwise comparisons between the Human network and each of the large LLM networks (Mistral, Llama3, and Haiku). The node comparison is straightforward, while the edge comparison considers only edges between common nodes of the graphs being compared.}
\label{tab:net_comp}
\end{table}


\section*{Data Records}

The LWOW datasets generated by Mistral, Llama3, and Haiku are accessible in the Github repository at the following link: \textit{https://github.com/LLMWorldOfWords/LWOW}. Each LLM dataset is provided as a .csv file that follows the structure shown in Table \ref{tab:csv_structure}. All data sources and code needed to reproduce the datasets and conduct further analyses are either available in the repository, or a description of where to access additional data sources (e.g. the SWOW dataset) is provided.\footnote{Due to the license of the SWOW dataset which prohibits the distribution of modified material, we do not provide a .csv file of the Human processed dataset, however, it can be generated using the code and other data sources provided in the repository.}

\begin{table}[h!]
\centering
\begin{tabular}{cccc}
\hline
\textbf{cue} & \textbf{R1} & \textbf{R2} & \textbf{R3} \\ \hline
apple        & banana      & fruit       & orange      \\
tree         & wood        & green       & leaf        \\
school        & teacher        & class       & building        \\
mathematics         & anxiety       & formula       & equation        \\
car         & train       & gasoline       & fuel        \\
...          & ...         & ...         & ...         \\ \hline
\end{tabular}

\caption{\textbf{Example dataset.} The table shows the structure of the .csv files. Each row has a cue word and three responses.}
\label{tab:csv_structure}
\end{table} 


\section*{Technical Validation}

To demonstrate the reliability of our data, we adopted a previously applied approach \cite{siew2019spreadr} that simulates the cognitive mechanisms underlying semantic priming, a cognitive phenomenon that entails recognizing target words more quickly when they are preceded by related prime words. Studies of semantic priming effects have been critical to gaining a better understanding of the nature of semantic memory \cite{hutchison2013semantic,collins1975spreading}. Semantic priming is usually investigated using the lexical decision task (LDT), in which human participants are presented with a prime word followed by a target word, and participants must decide as quickly as possible whether the target word is a real English word or a non-word. It has been found that participants identify the target word more quickly (lower reaction time) when the prime is related to the target (\textit{e.g. \textit{doctor} -- \textit{nurse}}) compared to when the prime is unrelated to the target (\textit{e.g. \textit{doctrine} -- \textit{nurse}}) \cite{neely2012semantic, mcnamara2005semantic}. This pattern has been shown to be consistent across thousands of prime-target pairs \cite{hutchison2013semantic}.

\begin{figure}[H]
\centering
\includegraphics[width=.7\linewidth]{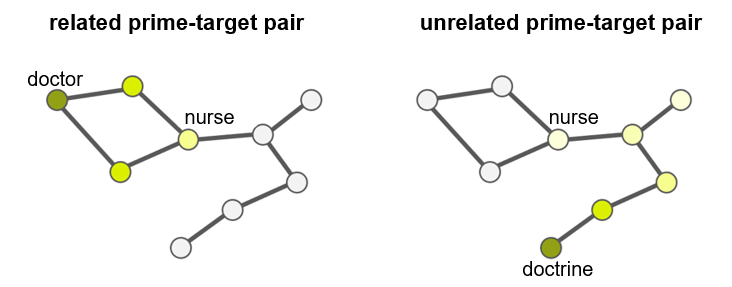}
\caption{\textbf{Spreading activation within a network.} The diagram shows how activation spreads within a network after various time steps after activating a prime node, leading to the activation of the target node \textit{nurse}. Darker colors indicate greater activation levels. On the left, the related prime \textit{doctor} is activated, while on the right, the unrelated prime \textit{doctrine} is activated. The final activation level of the target node \textit{nurse} is greater on the left when the activated prime is a related word.}
\label{fig:spr}
\end{figure}

In addition to the LDT, semantic priming can also be studied by implementing a spreading activation process within a network of semantic memory. Spreading activation is a method of search within a network that is based on supposed mechanisms of human memory \cite{collins1969retrieval, collins1975spreading}. In spreading activation theory, exposure to a concept leads to the activation of its corresponding node in semantic memory. That activation then propagates through the semantic memory network along the connections of the activated node, decaying over time, leading to the activation of other concepts in the network. This theory can be used to study semantic priming by simulating a search process within a network in which a prime node is activated and at the end of the spreading activation process, the  final activation level of the target node is observed. Following the semantic priming effect, the final activation level of the target node (e.g. \textit{nurse}) should, in theory, be greater when the activated prime is a related word (e.g. \textit{doctor}) rather than an unrelated word (e.g. \textit{doctrine}) \cite{siew2019spreadr}. This process is represented in Figure \ref{fig:spr}.

Spreading activation processes can be simulated within empirical networks using the R library \textit{spreadr} \cite{siew2019spreadr}. Given an empirical network, the \textit{spreadr} algorithm works by specifying one or more nodes in the network to be activated, the initial activation level of the activated nodes, and the number of iterations or time steps. At each time step, an activated node may retain a certain percentage of its activation, while the remaining percentage is distributed among its neighboring nodes, proportional to the weight of the edges if the network is weighted. This process continues iteratively until the specified number of time steps is reached. At the end of the iterative process, the final activation level of each node in the network can be measured.

The developers of \textit{spreadr} \cite{siew2019spreadr} simulated spreading activation within a free association network of semantic memory \cite{nelson2004university} to investigate the semantic priming effect using a series of prime-target pairs and their corresponding empirical reaction times from a lexical decision task experiment \cite{hutchison2013semantic}. As expected, they found that the final activation levels of the target nodes correlated with reaction times from the empirical data. That is, the final activation levels of the targets were greater when the activated primes were related rather than unrelated to the targets. These results show that the semantic priming effect observed in the psycholinguistic LDT experiment can also be observed within a network model of semantic memory, demonstrating the usefulness of semantic networks for modeling certain cognitive phenomena.

\begin{figure}[H]
\centering
\includegraphics[width=.7\linewidth]{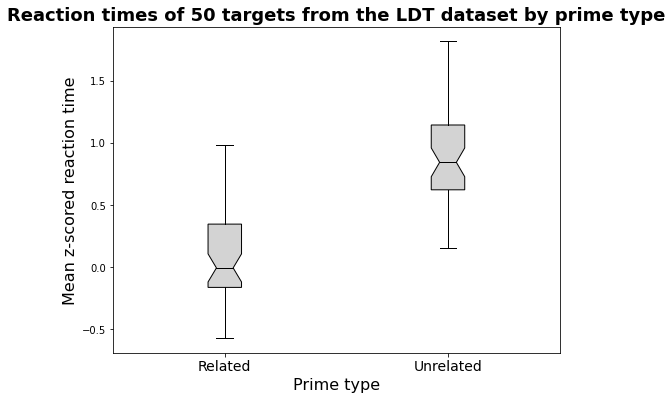}
\caption{\textbf{Differences in RTs from the LDT dataset.} The boxplot shows the difference in mean z-scored reaction times from the LDT experiment for related prime-target pairs and unrelated prime-target pairs. Reaction times for related prime-target pairs are, on average, less than those for unrelated prime-target pairs.}
\label{fig:box_RT}
\end{figure}

In order to validate our datasets, we repeated this spreading activation investigation of semantic priming implemented by the authors of \textit{spreadr}, but we used the Human and LLM networks that we built. We used a subset of 50 prime-target pairs and their corresponding reaction times from the same LDT dataset used by the authors of \textit{spreadr} \cite{siew2019spreadr}, downloadable from \textit{https://www.montana.edu/attmemlab/spp.html} \cite{hutchison2013semantic}. The 50 prime-target pairs and subsequent standardized reaction times (RTs) are shown in Table \ref{tab:LDT_primes_tgts} in the Appendix. Reaction times for the related prime-target pairs are, on average, less than those for the unrelated prime-target pairs. This pattern is shown in the boxplot in Figure \ref{fig:box_RT}. The significance of these paired differences (RTs of related prime-target pairs minus RTs of corresponding unrelated prime-target pairs) is confirmed by a Wilcoxon rank test for paired samples with an effect size of -0.87 (p < 0.001).

\begin{figure}[H]
\centering
\includegraphics[width=.8\linewidth]{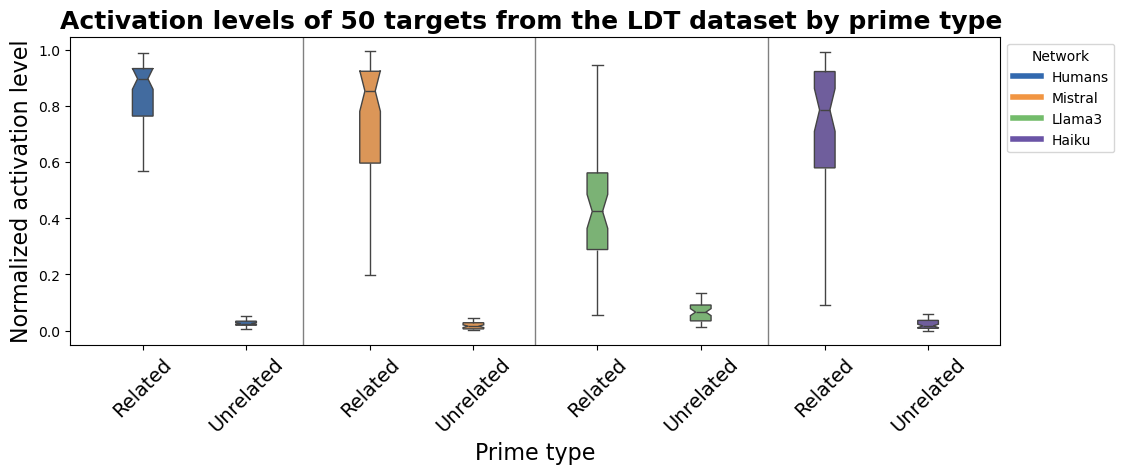}
\caption{\textbf{Validation of networks.} The boxplots show the differences in normalized final activation levels of the targets when related primes are activated compared to unrelated primes, for all networks. There is a significant difference in the final activation level, with higher final activation levels of the targets when the prime is related to the target for all networks.}
\label{fig:box_LDT}

\end{figure}

In a series of empirical simulations using \textit{spreadr} \cite{siew2019spreadr}, we activated each prime from all 100 prime-target pairs and we observed the final activation levels of the 50 corresponding targets. We repeated these simulations in each of the four networks. As previously discussed, the \textit{spreadr} library requires specification of the initial activation level of the activated node(s), and the number of iterations of the spreading activation process. We set the initial activation level of the prime node to the number of nodes in the entire network, and we set the number of iterations to two times the diameter of the network \cite{samuel2023predicting}. We specified that the networks are weighted so the edge weights were taken into consideration. Other parameters of the algorithm, such as the percentage of activation retained by each node, were set to default settings. For each network, the series of spreading activation processes yielded a matrix of final activation levels, such that the columns are all 100 prime nodes while the rows are all nodes of the network. Thus, each column of the matrix gives a vector of final activation levels of all nodes in the network after activating a single prime node, and a single entry in that vector is final activation level of the target node. The matrices were normalized first by normalizing columns of the matrix and then the rows. The normalization is necessary because it accounts for differences in the centrality of nodes within the semantic networks. By controlling for this factor, the normalized final activation levels reflect the semantic priming effects, and not effects related to node centrality, like frequency effects. We observed the normalized final activation levels of the target nodes when they were activated by related primes compared to when they were activated by unrelated primes, and we found results consistent with those obtained by the authors of \textit{spreadr} for all four networks. That is, final activation levels of targets are higher when they are activated by related primes compared to unrelated primes. Wilcoxon rank tests for paired samples confirm the statistical significance of these paired differences (all p < 0.001). These differences in activation levels by prime type are shown in the boxplots in Figure \ref{fig:box_LDT}. Effect sizes of these differences, which are all relatively large and comparable to the effect size of the differences in empirical RTs from the LDT experiment (-0.87), are shown in Table \ref{tab:effect_cor_LDT}. Similar to the authors of \textit{spreadr}, we also found that the normalized final activation levels that we observed correlated with the empirical RTs from the LDT experiment for all four networks. That is, higher activation levels are associated with lower reaction times. The Spearman correlations are shown in Table \ref{tab:effect_cor_LDT} (all p < 0.001). These results show that the network models of semantic memory built from the Human and LWOW datasets can be used for investigating semantic priming effects not only in humans but also in LLMs, demonstrating the validity of the datasets.

\begin{table}[H]
\centering
\begin{tabular}{lcccc}
\hline
\textbf{Effect size (prime type)}                         & \textbf{Humans}      & \textbf{Mistral}     & \textbf{Llama3}      & \textbf{Haiku}       \\ \hline
\multicolumn{1}{c}{}                                      & 0.870***               & 0.859***                & 0.869***                & 0.866***                \\
                                                          & \multicolumn{1}{l}{} & \multicolumn{1}{l}{} & \multicolumn{1}{l}{} & \multicolumn{1}{l}{} \\ \hline
\textbf{Correlation (activation level vs. reaction time)} & \textbf{Humans}      & \textbf{Mistral}     & \textbf{Llama3}      & \textbf{Haiku}       \\ \hline
\multicolumn{1}{c}{}                                      & -0.626***               & -0.615***               & -0.614***               & -0.662***               \\ \hline
\end{tabular}
\caption{\textbf{Effect sizes and correlations.} Effect sizes (female-related primes minus male-related primes) of the Wilcoxon rank test for paired samples are provided in the upper table. The Spearman correlation coefficient for activation levels vs. reaction times for each prime-target pair are provided in the bottom table (*** p < 0.001).}
\label{tab:effect_cor_LDT}
\end{table}




\section*{Usage Notes}

In the previous Section, we demonstrated the validity of the datasets by showing how they can be used to investigate semantic priming effects by simulating spreading activation processes within the networks. In this Section, we show how this methodology can be applied for investigating implicit biases within the networks, specifically, gender stereotypes. Since the semantic priming effect emerges due to the relatedness of words, observing this effect can be an indication that two words are related. Thus, semantic priming can be used to assess the strength of association between two words, and so, it can be implemented as a method for evaluating implicit biases, as discussed earlier \cite{schnabel2013free}. In what follows, we show how we can apply this theory and methodology to measure levels of gender biases within the networks.

\begin{figure}[H]
\centering
\includegraphics[width=.9\linewidth]{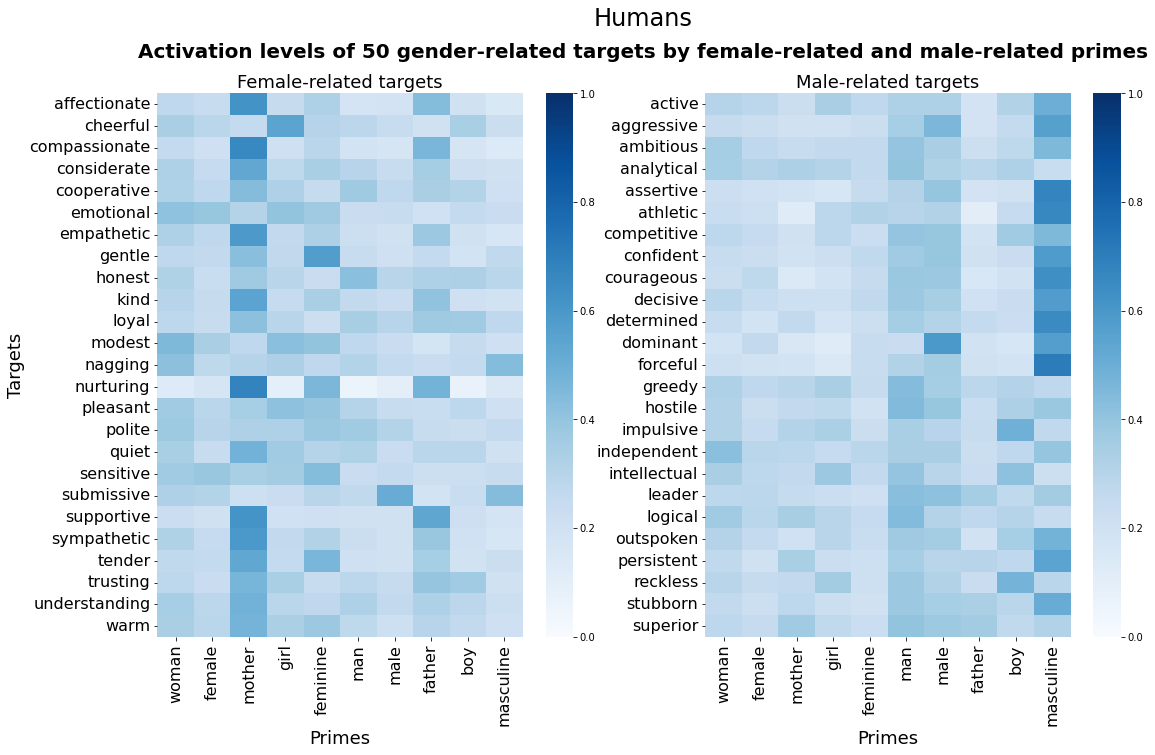}
\caption{\textbf{Gender biases in Humans.} Normalized final activation levels of the gender-related targets for the Human network.}
\label{fig:heat_humans}
\end{figure}

We start by choosing 5 female-related prime words (\textit{woman, female, mother, girl, feminine}) and 5 corresponding male-related prime words (\textit{man, male, father, boy, masculine}). We then choose 25 stereotypical female-related adjectives (e.g. \textit{gentle, emotional}) and 25 stereotypical male-related adjectives (e.g. \textit{strong, dominant}) as targets. These adjectives, which are listed in Table \ref{tab:gender_primes_tgts} in the Appendix, were taken from a study investigating gendered wording in job postings \cite{gaucher2011evidence}. Each prime is paired with a target, resulting in 500 prime-target pairs, half of which are stereotype consistent (e.g. \textit{woman} -- \textit{emotional}, \textit{man} -- \textit{dominant}) while the other half are stereotype inconsistent (e.g. \textit{man} -- \textit{emotional}, \textit{woman} -- \textit{dominant}). We then repeated the spreading activation methodology described in the previous section, activating all the primes in all four networks using \textit{spreadr} with the same parameter specifications as described earlier. We obtained normalized final activation levels of all 50 targets for each of the 10 primes. These activation levels for Humans are displayed in the heatmap in Figure \ref{fig:heat_humans}, with female-related targets on the left and male-related targets on the right. The same heatmaps for the LLMs (Mistral, Llama3, and Haiku) can be found in Figures \ref{fig:heat_mistral}-\ref{fig:heat_haiku} in the Appendix. These heatmaps provide a better understanding of the types of gender biases that are present in the semantic networks. For example, the heatmap for Humans on the right (Figure \ref{fig:heat_humans}) appears to have a clear division down the center, with darker colors on the right. This indicates overall higher activation levels of male-related targets when male-related primes are activated compared to female-related primes. A similar pattern is somewhat observable in the LLM heatmaps as well. The heatmaps are also useful for uncovering gender biases related to specific terms. For example, in Figure \ref{fig:heat_humans}, a large difference in activation level can be observed in the male-related target \textit{forceful} when activated by the prime \textit{masculine} compared to its female-related counterpart \textit{feminine}. Also, certain primes like \textit{mother} and \textit{masculine} appear to result in higher activation levels of targets across the board, indicating that these terms may carry more bias within the semantic networks.

\begin{figure}[H]
\centering
\includegraphics[width=.8\linewidth]{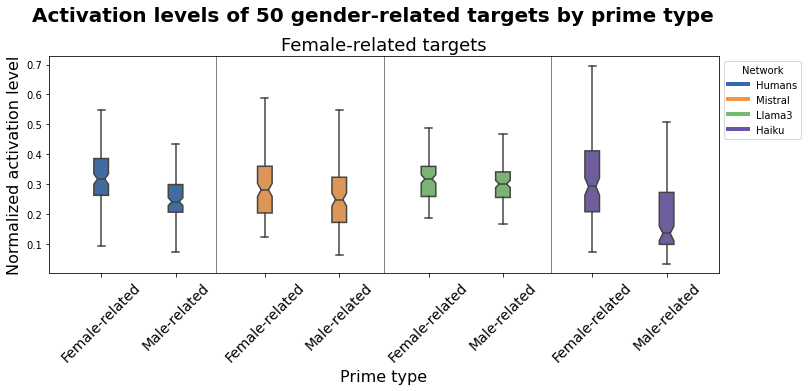}

\bigskip

\centering
\includegraphics[width=.8\linewidth]{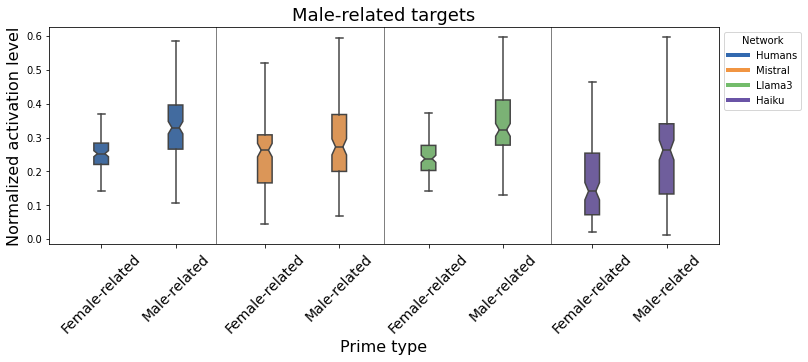}
\caption{\textbf{Validation.} The boxplots show the difference in normalized final activation levels of the targets when related primes are activated compared to unrelated primes, for all networks, for female-related targets (top) and male-related targets (bottom).}
\label{fig:box_gender}

\end{figure}

From the activation level matrices, we also calculated the paired differences of the activation levels (ALs) for all corresponding prime-target pairs (e.g. AL of \textit{forceful} when activated by \textit{feminine} minus AL of \textit{forceful} when activated by \textit{masculine}). We expected to observe higher levels of activation in stereotype consistent prime-target pairs (e.g. \textit{forceful -- masculine}) compared to stereotype inconsistent prime-target pairs (e.g. \textit{forceful -- feminine}). In fact, this is precisely what we found. We observed that among female-related targets, the normalized final activation levels were higher when activated by female-related primes rather than male-related, and vice versa among male-related targets. This pattern was observed in all four networks, but to varying degrees. These differences in activation levels are evidenced in the boxplots shown in Figure \ref{fig:box_gender}. Wilcoxon Rank tests for paired samples were done to assess the significance and the effect sizes of these paired differences. These effect sizes may be considered a measure of bias, since we would expect the difference in activation levels to be close to zero in the absence of bias. For example, in the absence of bias, we would expect both the female-related prime \textit{feminine} and its corresponding male-related prime \textit{masculine} to produce equal activation levels of the target \textit{forceful} (a difference close to zero), but instead we observed a large difference in activation level of this target in all four networks. This is an indication of the presence of bias. Thus, the effect size quantifies the extent to which this pattern exists (large differences in ALs caused by female-related primes vs. male-related primes), and thus, it quantifies the presence of gender bias in the networks. Effect sizes are shown in Table \ref{tab:effect_cor_gender}. For female-related targets, the largest effects (levels of gender bias) are observed for Humans and Haiku, with the smallest effects observed for Llama3. For male-related primes, the largest effects are observed for Humans and Llama3, with the smallest effects observed for Mistral. Histograms of paired differences in activation levels for each set of corresponding prime-target pairs are shown in Figure \ref{fig:hist_gender} in the Appendix. These histograms provide a clearer understanding of how the differences in activation levels differ from zero. Spearman correlations (all p < 0.001) between the final activation levels for Humans compared to each respective LLM are shown in Table \ref{tab:effect_cor_gender} as well. These correlations show how each LLM compares to Humans. We find that Llama3 differs the most from Humans for female-related primes, but that all other correlations are very similar (around 0.65).

This investigation demonstrates how the LWOW datasets presented in this paper can be used for investigating similarities and differences in the semantic capabilities of humans and LLMs, and in particular, for investigating the biases embedded in the conceptual knowledge structures of both humans and LLMs.

\begin{table}[H]
\centering
\begin{tabular}{lccccc}
\hline
\textbf{Effect size (prime type)}     & \textbf{Target type} & \textbf{Humans} & \textbf{Mistral} & \textbf{Llama3} & \textbf{Haiku} \\ \hline
\multicolumn{1}{c}{}                  & Female-related       & 0.686***           & 0.323***            & 0.153*           & 0.669***          \\
                                      & Male-related         & -0.695***          & -0.200*           & -0.722***          & -0.412***         \\
                                      &                      &                 &                  &                 &                \\ \hline
\textbf{Correlation (Humans vs. LLM)} & \textbf{Target type} & \textbf{Humans} & \textbf{Mistral} & \textbf{Llama3} & \textbf{Haiku} \\ \hline
\multicolumn{1}{c}{}                  & Female-related       & -               & 0.667***            & 0.334***           & 0.630***          \\
                                      & Male-related         & -               & 0.655***            & 0.648***           & 0.638***          \\ \hline
\end{tabular}
\caption{\textbf{Effect sizes and correlations.} Effect sizes (female-related primes minus male-related primes) of the Wilcoxon rank test for paired samples are provided in the upper table. The Spearman correlation coefficient for Human activation levels vs. LLM activation levels are provided in the bottom table (*** p < 0.001; * p < 0.05).}
\label{tab:effect_cor_gender}
\end{table}

\section*{Code availability}
The python and R code and all related data used to produce the LWOW datasets and conduct the analyses are available at \href{https://github.com/LLMWorldOfWords/LWOW}{https://github.com/LLMWorldOfWords/LWOW}.

\section*{Appendix}

\begin{table}[H]
\centering
\begin{tabular}{ccccc}
\hline
\textbf{Target} & \textbf{Related prime} & \textbf{Unrelated prime} & \textbf{Related RT} & \textbf{Unrelated RT} \\ \hline
britain         & england                & like                     & 0.75                & 2.21                  \\
dill            & pickle                 & cane                     & 0.21                & 1.56                  \\
coral           & reef                   & snob                     & 0.00                & 1.09                  \\
clipper         & toenail                & show                     & 0.05                & 1.11                  \\
christ          & jesus                  & chunk                    & -0.09               & 0.97                  \\
goo             & slime                  & anxiety                  & 0.98                & 2.02                  \\
communism       & fascism                & chicken                  & 0.47                & 1.48                  \\
ape             & chimpanzee             & foe                      & -0.35               & 0.64                  \\
id              & ego                    & obsession                & 0.70                & 1.66                  \\
novice          & beginner               & rod                      & 0.13                & 1.09                  \\
dive            & scuba                  & selfish                  & -0.39               & 0.56                  \\
clam            & oyster                 & sack                     & -0.09               & 0.85                  \\
extrovert       & introvert              & saying                   & 0.91                & 1.82                  \\
george          & curious                & honey                    & 0.17                & 1.06                  \\
doe             & deer                   & spontaneous              & 0.47                & 1.35                  \\
cereal          & breakfast              & official                 & -0.22               & 0.65                  \\
writer          & editor                 & zit                      & -0.33               & 0.53                  \\
murderer        & victim                 & sample                   & -0.01               & 0.84                  \\
weave           & basket                 & festival                 & -0.13               & 0.71                  \\
neutron         & atom                   & high                     & 0.38                & 1.21                  \\
immature        & juvenile               & survey                   & 0.14                & 0.97                  \\
reef            & coral                  & squash                   & -0.04               & 0.78                  \\
orgasm          & climax                 & snack                    & 0.34                & 1.15                  \\
shove           & push                   & kleenex                  & -0.11               & 0.70                  \\
assistant       & helper                 & obscure                  & -0.16               & 0.63                  \\
toast           & muffin                 & escargot                 & -0.49               & 0.30                  \\
pudding         & custard                & gold                     & -0.16               & 0.63                  \\
monk            & monastery              & opposite                 & 0.08                & 0.85                  \\
hula            & hoop                   & sister                   & 0.77                & 1.54                  \\
soprano         & alto                   & pardon                   & 0.52                & 1.28                  \\
decay           & decompose              & fairytale                & 0.09                & 0.85                  \\
reel            & fishing                & strive                   & 0.35                & 1.10                  \\
catholic        & methodist              & study                    & -0.12               & 0.63                  \\
fraternity      & greek                  & justify                  & 0.50                & 1.23                  \\
broccoli        & cauliflower            & metric                   & 0.04                & 0.77                  \\
tile            & roof                   & friendship               & -0.12               & 0.60                  \\
violin          & orchestra              & intestine                & -0.13               & 0.59                  \\
write           & essay                  & oxygen                   & -0.57               & 0.15                  \\
defeat          & conquer                & ballet                   & -0.07               & 0.65                  \\
lightning       & thunder                & hurt                     & -0.26               & 0.45                  \\
prince          & princess               & production               & -0.26               & 0.46                  \\
salad           & potato                 & car                      & -0.18               & 0.53                  \\
armor           & knight                 & ahead                    & 0.11                & 0.81                  \\
wimp            & coward                 & circle                   & 0.42                & 1.12                  \\
shoelace        & lace                   & rain                     & 0.52                & 1.21                  \\
rocket          & missile                & office                   & -0.36               & 0.33                  \\
store           & grocery                & white                    & -0.45               & 0.25                  \\
narrow          & straight               & strong                   & -0.16               & 0.53                  \\
am              & be                     & alarm                    & 0.25                & 0.93                  \\
stiff           & rigid                  & pudding                  & -0.06               & 0.62                  \\ \hline
\end{tabular}
\caption{\textbf{LDT dataset.} Set of 50 targets with related and unrelated primes and corresponding mean z-scored reaction times (RTs) from the lexical decision task dataset \cite{hutchison2013semantic}, used for data validation.}
\label{tab:LDT_primes_tgts}
\end{table}

\begin{table}[t!]
\centering
\begin{tabular}{cc}
\hline
\textbf{Female-related targets} & \textbf{Male-related targets} \\ \hline
affectionate                    & active                        \\
cheerful                        & aggressive                    \\
compassionate                   & ambitious                     \\
considerate                     & analytical                    \\
cooperative                     & assertive                     \\
emotional                       & athletic                      \\
empathetic                      & competitive                   \\
gentle                          & confident                     \\
honest                          & courageous                    \\
kind                            & decisive                      \\
loyal                           & determined                    \\
modest                          & dominant                      \\
nagging                         & forceful                      \\
nurturing                       & greedy                        \\
pleasant                        & hostile                       \\
polite                          & impulsive                     \\
quiet                           & independent                   \\
sensitive                       & intellectual                  \\
submissive                      & leader                        \\
supportive                      & logical                       \\
sympathetic                     & outspoken                     \\
tender                          & persistent                    \\
trusting                        & reckless                      \\
understanding                   & stubborn                      \\
warm                            & superior                      \\ \hline
\end{tabular}
\caption{\textbf{Gender-related targets}. Female-related and male-related targets used to investigate gender biases within the networks, taken from a previous study investigating gendered language in job postings \cite{gaucher2011evidence}.}
\label{tab:gender_primes_tgts}
\end{table}

\begin{figure}[H]
\centering
\includegraphics[width=.9\linewidth]{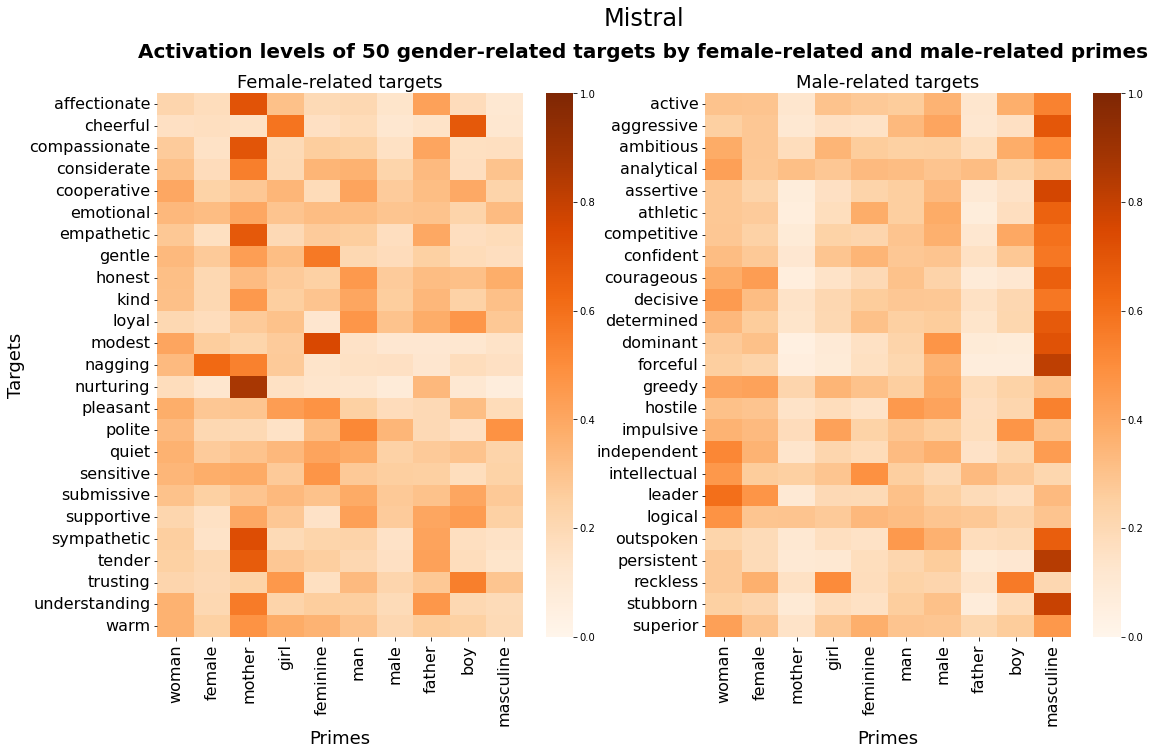}
\caption{\textbf{Gender biases in Mistral.} Normalized final activation levels of the gender-related targets for the Mistral network.}
\label{fig:heat_mistral}
\end{figure}

\begin{figure}[H]
\centering
\includegraphics[width=.9\linewidth]{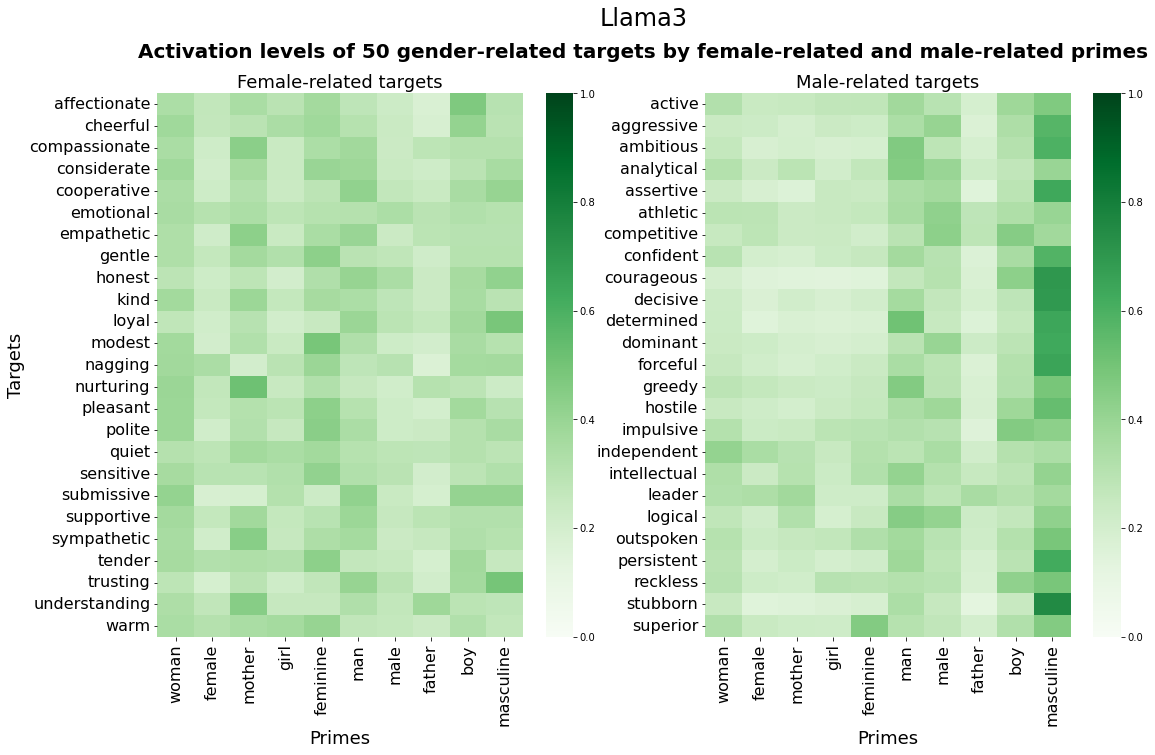}
\caption{\textbf{Gender biases in Llama3.} Normalized final activation levels of the gender-related targets for the Llama3 network.}
\label{fig:heat_llama3}
\end{figure}

\begin{figure}[H]
\centering
\includegraphics[width=.9\linewidth]{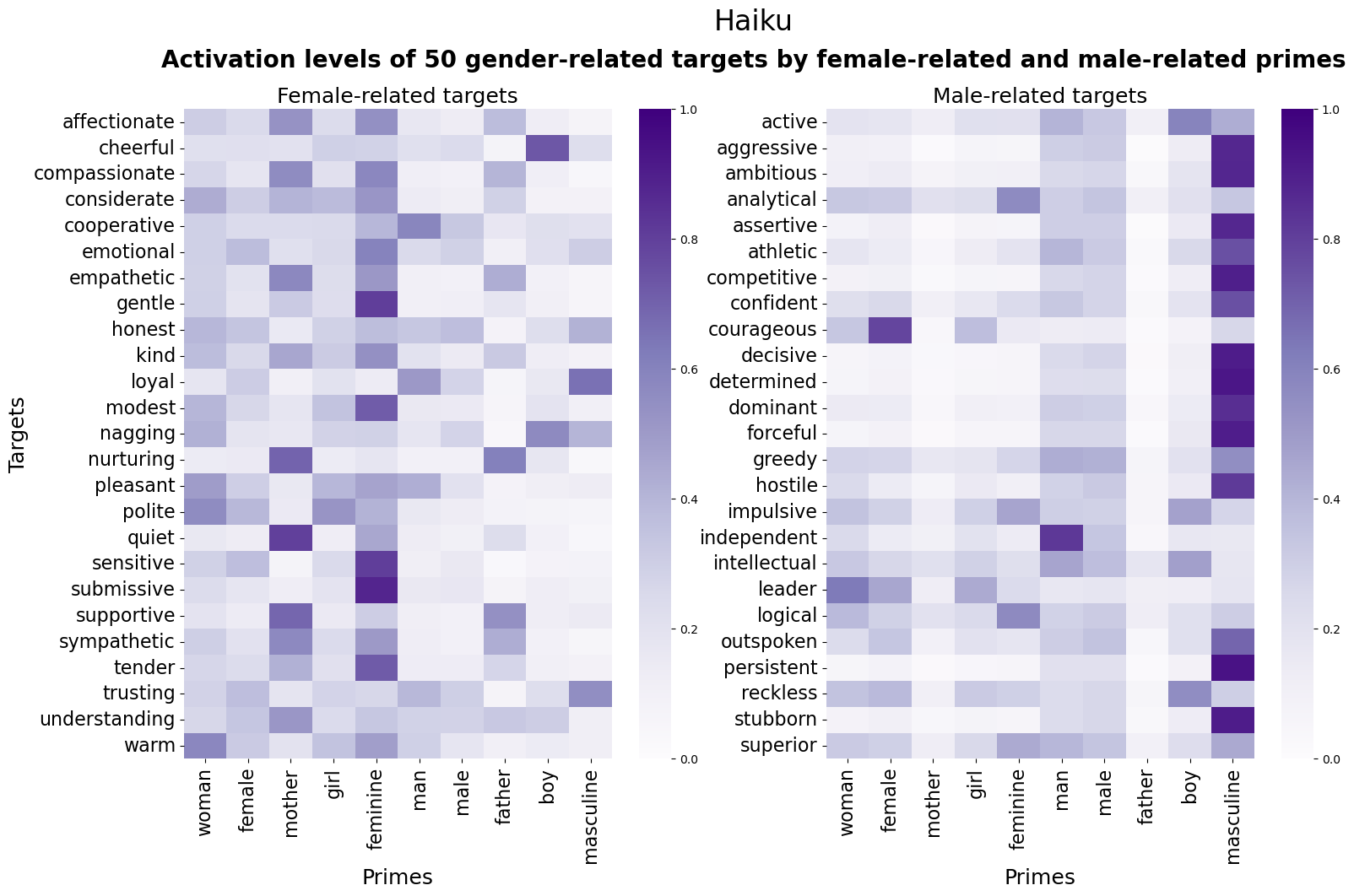}
\caption{\textbf{Gender biases in Haiku.} Normalized final activation levels of the gender-related targets for the Haiku network.}
\label{fig:heat_haiku}
\end{figure}

\begin{figure}[t!]
\centering
\includegraphics[width=.8\linewidth]{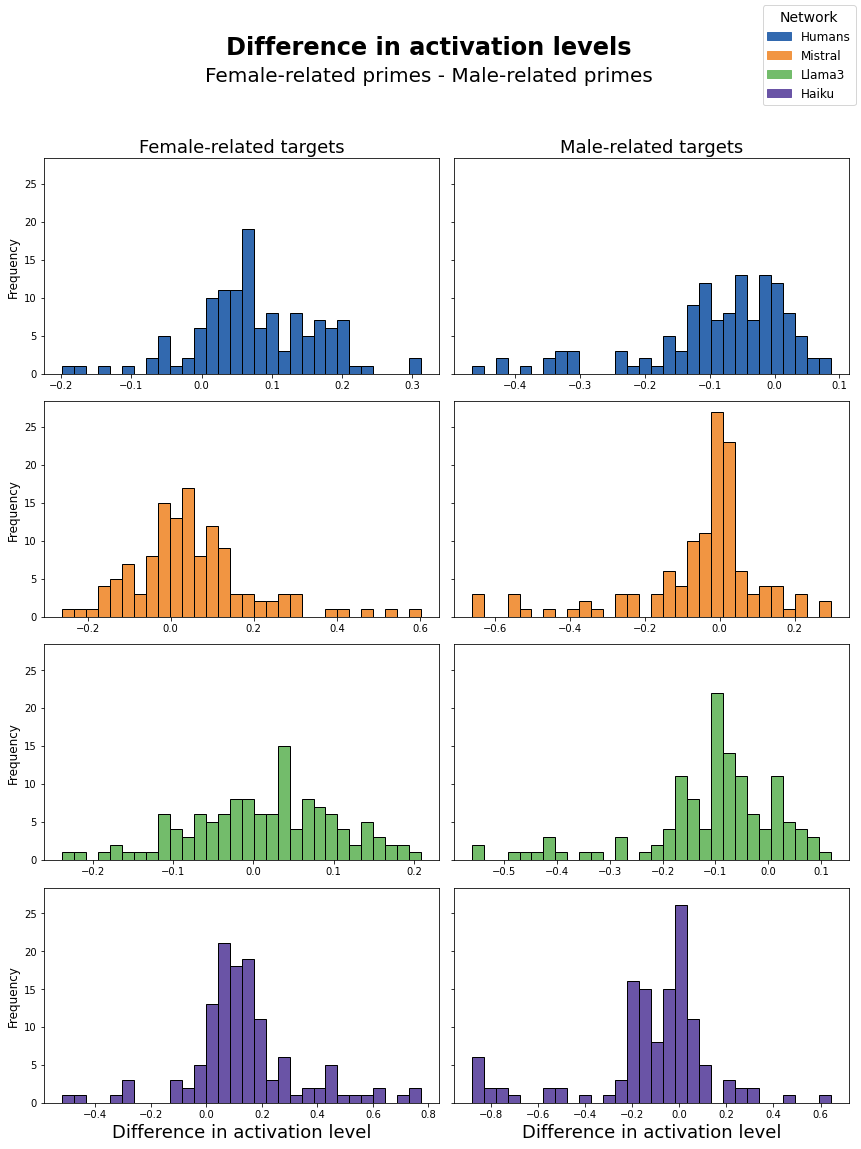}
\caption{\textbf{Distributions of differences in activation levels.} Histograms of paired differences in activation levels for each prime-target pair. These histograms provide a clearer understanding of how the differences in activation levels differ from zero.}
\label{fig:hist_gender}
\end{figure}


\bibliography{main}



\section*{Author contributions statement}

K.A., G.R. and M.S. conceived the experiments, K.A., R.I. and G.R. conducted the experiments, K.A. analyzed the results and visualized the data. All authors reviewed the manuscript.


\section*{Competing interests} 
The authors declare no competing interests.

\end{document}